\documentclass[fleqn]{article}
\usepackage{nips13submit_e2,times}
\usepackage{url}
\usepackage{graphicx}
\usepackage{cite}
\usepackage{amsmath}
\usepackage{amssymb}
\usepackage{color}
\usepackage{bm}
\usepackage{footnote}
\usepackage{verbatim}


\begin{document}

\title{\LARGE{A new model for Cerebellar computation}}
\author{
\large{\textbf{Reza Moazzezi \textsuperscript }} \\ 
 \textsuperscript{}
{r.moazzezi@gmail.com} 
\small{}
}

\nipsfinalcopy

\maketitle

\section{Abstract}
\label{sec:abstract}

The standard state space model is widely believed to account for the cerebellar computation in motor adaptation tasks \cite{shadmehr2005computational}. Here we show that several recent experiments \cite{vaswani2015persistent, morehead2017characteristics, kim2017invariant} where the visual feedback is irrelevant to the motor response challenge the standard model. Furthermore, we propose a new model that accounts for the results presented in \cite{vaswani2015persistent, morehead2017characteristics, kim2017invariant}. According to this new model, learning and forgetting are coupled and are error size dependent. We also show that under reasonable assumptions, our proposed model is the only model that accounts for both the classical adaptation paradigm as well as the recent experiments \cite{vaswani2015persistent, morehead2017characteristics, kim2017invariant}. 

\section{Introduction}
\label{sec:intro}

\subsection{Standard state space model}

In Visuo-Motor Rotation experiments (VMR), subjects are instructed to land a cursor on a target. At some point during the experiment, the visual feedback is rotated and the motor system's reaction (in the form of motor adaptation) is measured in response to the rotation. The standard state space model (standard SSM) is shown to account for the VMR experiment. Standard SSM has the following form:

\begin{equation}
\label{eq:ssm0}
\begin{split}
X_{n+1}=A X_n + B E_n
\end{split}
\end{equation}

where $X_n$ is the cursor location on trial $n$, $E_n$ is the error at trial $n$, and $A$ and $B$ are constants. Error $E_n$ on trial $n$ is measured as the difference between the target location and the cursor location, that is, $E_n=T-X_n$ where $T$ is the target location and $X_n$ is the cursor location on trial $n$. As mentioned above, Standard SSM accounts for the adaptation observed in VMR. 

\subsection{Task Irrelevant Clamped Visual Feedback experiments}
\label{subsec:TICVF}

In a series of recent experiments \cite{vaswani2015persistent, morehead2017characteristics, kim2017invariant}, unlike VMR, the visual feedback (that is a measure of the error made by the subjects) was clamped and subjects were asked to move a cursor towards a target (aim at a target) and \textbf{ignore} the visual feedback. Since the visual feedback was clamped and therefore was irrelevant to the task, we refer to this task as Task Irrelevant Clamped Visual Feedback paradigm (\textbf{TICVF}). 

TICVF data suggests 3 key features:

\begin{enumerate}

\item The saturation level of adaptation is independent of the error size for a broad range of errors

\item The initial rate of adaptation depends on error size for errors less than 7.5 degrees

\item The initial rate of adaptation is independent of error size for errors larger than 7.5 degrees
\end{enumerate}

In this paper, we show that these key features of TICVF are inconsistent with the predictions of the standard SSM.

In addition, we propose that the following model accounts for the key features of the TICVF paradigm:

\begin{equation}
\label{eq:newmodel}
\begin{split}
X_{n+1}=(1-P_{(E_n)})X_n+P_{(E_n)}K
\end{split}
\end{equation}\\

where $X_n$ is cursor position on trial $n$, $K=+k$ if the sign of error is positive, $K=-k$ if the sign of error is negative and $K=0$ if error is zero, $E_n$ is the error on trial $n$ (error is measured in degrees) and $P_{(E)}$ is a nonlinear function of $E$ (for example, $P_{(E)}$ is a sigmoid function, $P_{(E)}=\frac{a}{b+exp(-cE)}$ or a linear approximation to that, which saturates at about 7.5 degrees).  

\section{Results}

\subsection{TICVF paradigm and our model}
\label{subsec:TICVFmodel}

As we discussed earlier, in TICVF paradigm, visual feedback is irrelevant and the error is clamped and constant throughout the whole experiment. In addition, subjects are instructed to aim at the target and ignore the visual feedback. For the TICVF paradigm equation \ref{eq:newmodel} is simplified as follows :

\begin{equation}
\label{eq:ticvf}
\begin{split}
X_{n+1}=(1-P_{(E)})X_n+P_{(E)}K
\end{split}
\end{equation}\\

where $E$ is the clamped error (and is fixed during the experiment). 

\subsection{Standard model fails to account for TICVF key features}
\label{subsec:standard}

The standard model in TICVF paradigm takes the following form:

\begin{equation}
\label{eq:ssm1}
\begin{split}
X_{n+1}=A X_n + B E
\end{split}
\end{equation}

where $E$ is the clamped error. As $n \rightarrow \infty$, $X_n$ converges, that is, $X_{n+1}=X_n  \triangleq X$. At the point of convergence we have:

\begin{equation}
\label{eq:ssm2}
\begin{split}
X=A X + B E \Rightarrow X=\frac{B E}{1-A}
\end{split}
\end{equation}

This implies that if the error $E$ is doubled, then the saturation level, $X$, is also going to double, which is at odds with the first key feature of TICVF experiment (according to which saturation level is independent of the error $E$).

In addition, without loss of generality, we can assume that $X_0=0$. As a result, based on equation \ref{eq:ssm1}, $X_1=B E$. If we think of $X_1-X_0$ as a proxy for initial slope (note that $X_1-X_0=X_1$ because $X_0=0$), then it is clear that the inital slope is proportional to $E$ for all error sizes, while TICVF experiment suggests that the initial slope is independent of $E$ for $E>7.5$ degrees. 

As a result, it is clear that the standard model fails to account for the key features of the TICVF paradigm.

\subsection{Our model accounts for the three key features of the TICVF experiment}
\label{subsec:key}

In this section we show that our model accounts for all the key features of the Error Clamp experiment described in section \ref{subsec:TICVF}:

\begin{enumerate}

\item  we first show that as $n$ goes to infinity, $X_n$ asymptotes at $K$ and this is independent of the error size (as long as $E > 0$. We assume that $P_{(0)}=0$ and if $E>0$ then $P_{(E)} > 0$):

Note that as $n \rightarrow \infty$, we have $X_{n+1}=X_n \triangleq X$. Given that in TICVF, $E_n=E$ for all $n$, as long as $P_{(E)} \neq 0$, we have (based on equation \ref{eq:ticvf}):

\begin{equation}
\label{eq:k}
\begin{split}
X=(1-P_{(E)})X+P_{(E)}K \Rightarrow X=K
\end{split}
\end{equation}\\

In other words, the final adaptation level, $X$, is $K$ and is independent of the error size, $E$.

\item Since $P_{(E)}$ is constant for $E > 7.5$ degrees, for $E > 7.5$, similar to TICVF, the learning curve (and in particular the initial slope) is independent of the error size. Let's denote $ P_{(E)}$ for $E > 7.5$ degrees by $1-A$, that is, $P_{E>7.5}=1-A$. Then:

\begin{equation}
\label{eq:largeE}
\begin{split}
X_{n+1}=AX_n+ (1-A) K \ \ \ \ \ \ \ \ for \ \ \ \ \ E > 7.5
\end{split}
\end{equation}\\

Therefore, for $E > 7.5$, the initial slope is independent of the error size.

\item for $E < 7.5$ degrees, $P_{(E)}$ depends on $E$ (and is an increasing function of $E$) and therefore, similar to TICVF, the initial slope also scales with $E$. 

\end{enumerate}

\subsection{Assumptions under which our model is the only model that accounts for TICVF results}

Are there assumptions under which our model is the only model to account for the key features of TICVF? We are interested to model the key features of the TICVF experiment by an update rule similar to the standard state space model. In particular, we are interested in a model with the following properties:

\begin{enumerate}

\item State at time $n+1$, $X_{n+1}$, depends only on $X_n$ (similar to the standard model)

\item $X_{n+1}$ depends linearly on $X_{n}$ (similar to the standard model)

\item $X_{n+1}$ only depends on the error on trial $n$

\end{enumerate}

In other words, we constrain the model to the following form:

\begin{equation}
\label{eq:general-linear0}
\begin{split}
X_{n+1}=f_{(E_n)} X_n + g_{(E_n)}
\end{split}
\end{equation}\\

where $f(.)$ and $g(.)$ are general fuctions (of error) and our goal is to find out under what conditions (if any) these functions account for the key features of TICVF. In TICVF, the above equation turns into the following equation:

\begin{equation}
\label{eq:general-linear}
\begin{split}
X_{n+1}=f_{(E)} X_n + g_{(E)}
\end{split}
\end{equation}\\

Without loss of generality, we can assume that:

\begin{equation}
\label{eq:general-lin-assumption}
\begin{split}
g_{(E)}= P_{(E)} K
\end{split}
\end{equation}\\

where $K$ is a constant (in other words, we define a new function $P_{(E)}=\frac {g_{(E)}}{K}$). We will show that, under the above three assumptions, the only way for the model to have the same asymptote for all error sizes (one of the key features of the TICVF experiment) is for $f_{(E)}$ to be equal to $f_{(E)}=1-P_{(E)}$. In other words, we will show that under assumptions 1 to 3, our model (equation \ref{eq:ticvf}) is the only possible solution. To prove this, note that by combining equations \ref{eq:general-linear} and \ref{eq:general-lin-assumption}, we have:

\begin{equation}
\label{eq:general-linear1}
\begin{split}
X_{n+1}=f_{(E)} X_n + P_{(E)} K
\end{split}
\end{equation}\\

If we assume that as $n \rightarrow \infty$, $X_n \rightarrow K$, then in the limit we have:

\begin{equation}
\label{eq:general-linear-result}
\begin{split}
K=f_{(E)} K + P_{(E)} K \Rightarrow f_{(E)} + P_{(E)}=1 \Rightarrow f_{(E)}=1-P_{(E)}
\end{split}
\end{equation}\\

Combining \ref{eq:general-linear1} and \ref{eq:general-linear-result}, we have:

\begin{equation}
\label{eq:general-linear2}
\begin{split}
X_{n+1}=(1-P_{(E)}) X_n + P_{(E)} K
\end{split}
\end{equation}\\

which is the same equation as equation \ref{eq:ticvf}. Note that all the three assumptions mentioned above are basically the assumptions that also underly the standard SSM.

\subsection{Our model and the VMR experiment}
\label{VMR}

In the previous sections we have shown that the standard SSM accounts for VMR but fails to account for TICVF. We have also shown that our model accounts for TICVF. Here we show that our model accounts for VMR as long as the rotation in VMR is less than $K$ (see equation \ref{eq:newmodel}). To see this, note that in VMR, as the state ($X_n$) changes, error size ($E_n$) decreases. As $E_n$ decreases, $P_{(E_n)}$ starts to decrease when $E_n$ goes below 7.5 degrees ($P_{(E_n)}$ is a sigmoid that saturates around 7.5 degrees). Eventually, as $E_n$ goes to zero, $P_{(E_n)}$ also goes to zero and as a result, when $P_{(E_n)}=0$, $X_{n+1}=X_n$ in equation \ref{eq:newmodel}. Therefore, our model accounts for the VMR experiment. 

\section{Summary}

We proposed a new model for motor adaptation that accounts for the new TICVF results as well as the VMR experiment. Our model is radically different from the standard model. In the new model, both the learning rate and the forgetting rate are functions of the error size while in the standard model they are both independent of the error size. The standard model accounts for VMR but fails to account for TICVF. Our model accounts for both experiments. 

Our model also suggests that the learning rate and forgetting rate are coupled: For example, for large errors, the forgetting rate ($1-P_{(E)}$) is small and the learning term ($P_{(E)} K$) is large. Further experiments are needed to fine-tune, modify and improve our model. However given the current data (VMR + TICVF), it is clear that the model proposed in this paper is superior to the standard model as it explains both the VMR experiment as well as the TICVF.

\section*{}
\label{REF}
{\small
\bibliographystyle{unsrt}
\bibliography{REF}
}

\end{document}